\documentclass{research4cacm}
\usepackage{afterpage}
\usepackage{placeins}
\usepackage{enumitem}
\usepackage{float}
\usepackage{hyperref}
\hypersetup{
    colorlinks=true,
    linkcolor=blue,
    filecolor=magenta,      
    urlcolor=cyan,
}

\setenumerate{noitemsep}
\begin{document}
%
\CopyrightYear{2020} 
\title{Using LSTM and SARIMA Models to Forecast Cluster CPU Usage
%
}

%
%
%
%
%
\numberofauthors{2} 
%
\author{
%
%
\alignauthor
Langston Nashold\\
       \affaddr{Stanford University}\\
       \email{lnashold@stanford.edu}
\alignauthor
Rayan Krishnan\\
       \affaddr{Stanford University}\\
       \email{rayank@stanford.edu}
}

\maketitle
\begin{abstract}
   As large scale cloud computing centers become more popular than individual servers, predicting future resource demand need has become an important problem. Forecasting resource need allows public cloud providers to proactively allocate or deallocate resources for cloud services. This work seeks to predict one resource, CPU usage, over both a short term and long term time scale. 
   
   To gain insight into the model characteristics that best support specific tasks, we consider two vastly different architectures: the historically relevant SARIMA model and the more modern neural network, LSTM model. We apply these models to Azure data resampled to 20 minutes per data point with the goal of predicting usage over the next hour for the short-term task and for the next three days for the long-term task. The SARIMA model outperformed the LSTM for the long term prediction task, but performed poorer on the short term task. Furthermore, the LSTM model was more robust, whereas the SARIMA model relied on the data meeting certain assumptions about seasonality. 
   
   \medskip
\textit{The code for this paper can be found at}

   \textit{github.com/rayankrish/CpuUsagePrediction}
\end{abstract}


\section{Introduction}
\subsection{Motivation}
There has been a shift from smaller, individually-owned servers to massive datacenters with thousands of machines. One of the cloud computing providers, Amazon Web Services, has over a million active users \cite{VisualCapitalist}. Server racks can be split between multiple users, with each user having an isolated sandbox (a virtual machine) for their application. A user can even have an application running across multiple VMs on separate machines \cite{badrulhisam2019}. 

Cloud provider must chose how to allocate VMs to their different users. Users that have higher demand for resources (e.g. processing power, memory, etc.) should be allocated more virtual machines. Users with lower demand can be allocated less. There are two general approaches to VM allocation: reactive and proactive \cite{tahir_online_2020}. In a reactive approach, users are only allocated more resources once they begin to run out. In a proactive approach, the resource usage will be predicted ahead of time. 

Proactively allocating VMs can increase usage efficiency of underlying resources. If a computing cluster predicts the future resource usage of a user service will increase, it can preemptively scale up to accommodate a higher load. This allows a data center to meet its service level obligations (SLOs), which are agreements between cloud providers and customers to provide a certain quality of service. If it predicts that usage will decrease, it can deallocate VMs and save computing resources \cite{tahir_online_2020}. 

One very commonly studied resource in the context of VM allocation is CPU usage \cite{gupta_resource_2017}. Ideally, every processor in a data center should not be over or underutilized. This work pertains to predicting a future value of the CPU load (usually expressed as a number between 1 and 100). In the future, this research can inform additional models that make a decision to allocate or deallocate cloud machines based on company-specific cost-benefit analysis. Generally, VM allocation requires prediction on a relatively short time scale. 

However, there is also value in long term CPU usage predictions as well. Long term forecasts of resource use can be used to inform the purchase of equipment by IT professionals managing computing clusters \cite{kumar_forecasting_2016}. Currently, prediction is often based on unquantifiable industry expertise. Instead, machine learning techniques could be applied to more accurately predict hardware need \cite{long-term}. Therefore, there is a need for both long term and short term CPU usage forecasts. 

\begin{figure*}[t]
    \centering
    \caption{Sample data file plotted in blue. Data resampled and averaged to 20 points per point shown in orange.}
    \includegraphics[scale=0.4]{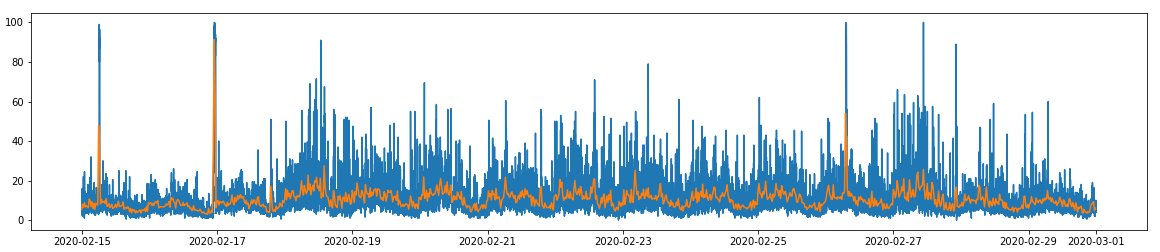} \\
\end{figure*}

\subsection{Prior Work}
    
    Historically, a wide variety of methods have been used to do classical time series analysis. One such approach involves simply computing an average over recent data points (an SMA model). Alternatively, exponential weighting can be used to provide a bias towards more recent data points (an EMA model)  Another approach, singular spectral analysis (SSA), involves decomposing a time series into its component parts using PCA, and then using the components to make future predictions. 
    
    Many of these methods have been used to predict resource usage for server clusters. Masdari and Khoshvenis have surveyed past methods used to predict CPU usage, which include the following classical techniques: \cite{masdari_survey_2019}.
    \begin{enumerate}[noitemsep]
        \item Wavelet transforms
        \item ARIMA models
        \item Markov Models \cite{zhenhuan_gong_press_2010}
        \item Support Vector Regression
        \item Singular Spectral Analysis
        \item Linear Regression
    \end{enumerate}
    
    Among these classical models, the ARIMA model has been notably very widely applied to both for timeseries in general and CPU forecasting specifically \cite{yang_workload_2013}. For example, Kumar and Mazumdar use several varieties of ARIMA model (SARIMA, ARMA, and SARFIMA) to predict resource usage inside high performance computing clusters \cite{kumar_forecasting_2016}. Other researchers have used ARIMA models as a baseline with which to compare more sophisticated techniques \cite{janardhanan_cpu_2017, gupta_resource_2017}. 
    
    Modern neural networks have also been applied to this problem. One of the most common deep learning models applied to time series is the long short term memory network, which overcomes the vanishing gradient problem associated with other neural networks. For example, Gupdta and Dinesh use a bidirectional long short term memory network (LSTM) to predict CPU usage \cite{gupta_resource_2017}. Janardhanan and Barrett also used an LSTM to provide CPU usage predictions \cite{janardhanan_cpu_2017}. Other models used have used a Deep Belief Network to generate predictions \cite{qiu_deep_2016}. 
    
    It is also important to note that these predictions are done on a wide variety of time scales. These range between a period of many months to only a minute \cite{kumar_forecasting_2016}. Different forecast lengths have different advantages. For example, a month-long prediction could be useful when predicting how many machines to fill your HPC cluster with. However, for VM auto-scaling, a much smaller horizon would be useful. 
    , 
\subsection{Contribution}   
We applied one state-of-the-art neural network, an LSTM, and one well-studied and robust classical model, the ARIMA model. We applied these models to a previously unstudied set of CPU load time series. By applying them to several disparate timeseries, we were able to compare both the overall performance of the two models with naive models and each other, as well as their strengths with regard to characteristics of the different data sets. This analysis of the strengths and weakness of each model can be applied in the evaluative phase of cluster resource time series analysis. This specific prediction of future usage can be applied to a more nuanced task of deciding on scaling resource allocation by an evaluation of risks.

\begin{figure}[!ht]
    \centering
    \caption{Distribution of data set means}
    \includegraphics[scale=0.25]{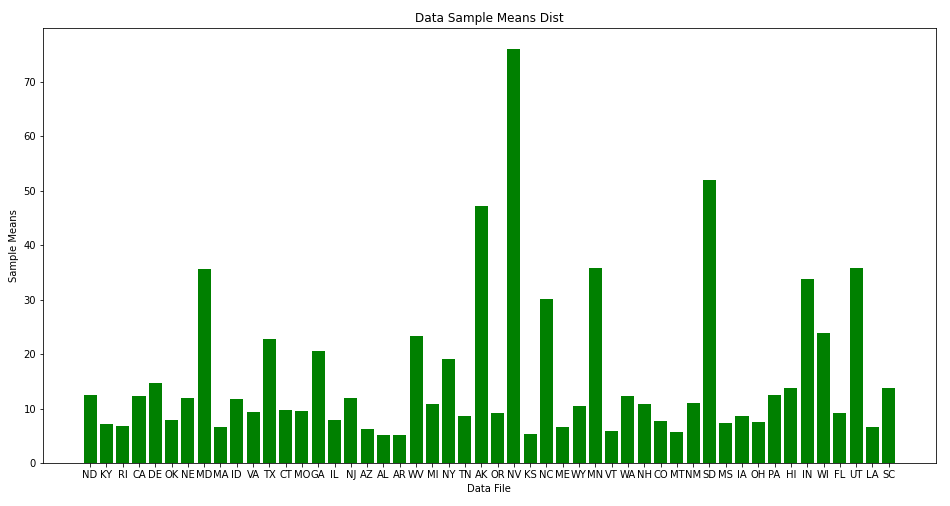} \\
\end{figure}

\begin{figure}[!ht]
    \centering
    \caption{Distribution of data set standard distributions}
    \includegraphics[scale=0.25]{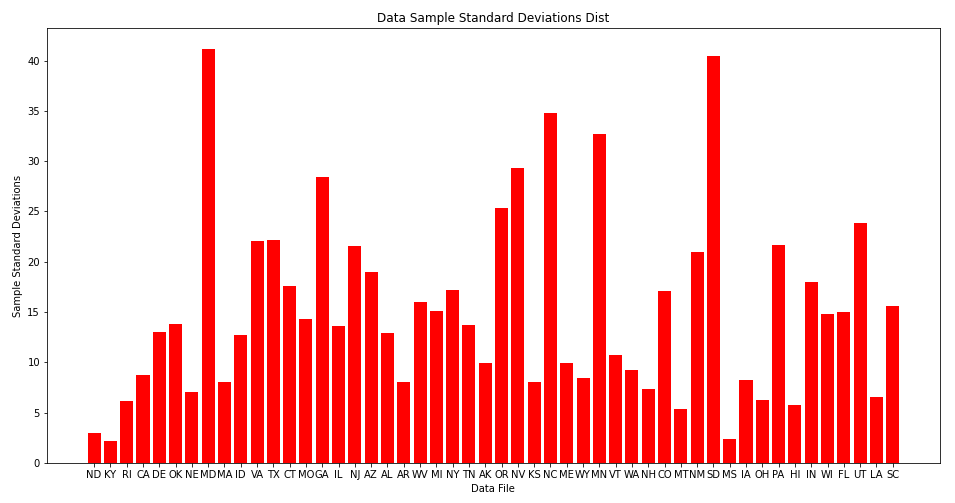}
\end{figure}
\FloatBarrier

\section{Data Analysis}
The data from which we base our work was obtained from a researcher at Microsoft Azure with his explicit permission. It has the data of 50 compute clusters in one-minute intervals over the course of two weeks. The data includes a set of samples as well as an identically sourced "holdout" set. Each data set is  labeled with a state abbreviation, with which they are referred to by in our analysis. These IDs are simply for the sake of distinguishing data sets and do not relate to the American states themselves. Each data set is a univariate time series over a 15 day period, with values ranging from 0 to 100 to represent CPU usage. To reduce noise and improve performance among models, this data is down sampled such that each point is 20 minutes apart. This down sampling is observed in Figure 1.

An initial data analysis shows that the individual clusters are non-identically distributed. As seen in the distribution of means (Figure 2), the data sets varied drastically. The mean of the means was 12.9, while the standard deviation of the means was 13.9. Further examination of the data leads us to believe that the cause of this variance is the sustained use or lack of use of clusters. Those clusters which exhibited infrequent use often reported CPU usage close to 0 while others with regular, sustained usage reported consistently high CPU usage values. This is shown in Figure 3. Those clusters which had middling average usage scores tended to also have higher variance than the others which were usually heavily or lightly used. As a result, the distribution of the standard deviations was not tightly clustered either. The standard deviation of the standard deviations of the data sets was 9.2. There are other topical trends we can observe, such as the increased cluster usage on weekdays over weekends.

We also inspected the individual cluster traces. First, we performed a decomposition to separate the time series into trend, seasonal, and noise components. An example of this is shown below, for the "WI" time series. \\

\begin{center}
\includegraphics[scale=0.6]{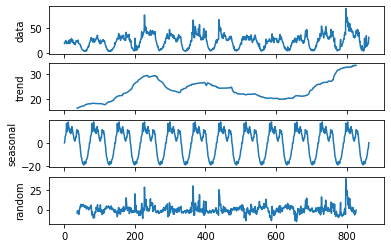}
\end{center}

Clearly, this data set has a strong daily seasonal component. In other words, since the data is seasonal, each data point is similar to the data points 24 hours ago. This makes sense, as one might expect there to be consistent daily patterns in the CPU usage, based on the users of the service and their daily habits. 

Furthermore, later models we use depend on the time series being stationary. To be stationary, a time series must have a constant variance and a constant mean. To test stationarity, the  Kwiatkowski–Phillips–Schmidt–Shin (KPSS) test was performed on every data sets. It was found that the vast majority of the data sets were stationary. It is possible that the two week window of data we analyzed was not indicative of a larger seasonality or trend of data that would have been captured in a year long data sample.
\section{Method}

\subsection{Baselines and Data Processing}
It is difficult to quantify the success of a model from it's error alone. We first implemented two naive models with which we can compare the error results. First, the naive prediction model simply predicts the last seen training value at each time. Second, the naive mean prediction predicts the mean of the previously seen training examples.

In the deployment of these models, our intent is to be able to predict CPU usage for two time horizons: over an hour and over three days (specifically, 20\% of the dataset). Predicting over the next hour allows a increase or decrease VM resources by a fixed amount. Predicting over several days could allow a cloud computing cluster owner to make longer term business decisions. 

As will be further described, the SARIMA model did not converge when being fit to 21600 data points. In order to allow for performant models, we first resampled the data to bins of 20 points each and averaged each bin. By nature, the SARIMA model fits the entire training data at once and predicts the test data. 

We implemented the LSTM with a data set produced through a "sliding window" method where by each input data is a vector with 6 data points for 2 hours of data and the model predicts 3 data points, or one hour of data. This imitates the real data a model might receive from continuous updates of usage. The model is both trained and tested on sliding window data.

\subsection{SARIMA Model} 
    The ARIMA model has been shown to have been effective on similar problems before \cite{janardhanan_cpu_2017}. Although the ARIMA is common used for time-series prediction, we implemented the SARIMA model to exploit underlying seasonality in our data. An implementation of both a classical model and a neural network offers the opportunity to consider how these underlying differences in architecture affect the accuracy and performance of each model type.
    
    The autoregressive integrated moving average (ARIMA) model is a combination of several simpler models, from which it gets its name \cite{Hyndman}. An autogressive model, denoted AR($p$), makes prediction for a future point $y_t$ based on the linear combination of the last $p$ points ($y_{t-1}$ ... $y_{t-p}$ and weights $\phi_i$: 
        \[ y_t = c + \phi_1 y_{t-1} + \phi_2 y_{t-2} + ... + \phi_{p} y_{t-p}\]
    A moving average, MA($q$), instead uses past forecast errors $\epsilon_t$ in regression, so 
        \[y_t = c + \theta_1 \epsilon_{t-1} + \theta_2 \epsilon_{t-2} +... + \theta_q \epsilon_{t-q}\] 
    An ARIMA model takes the sum of the autoregressive and moving average models. Furthermore, both the AR model and MA model assume the data is stationary. To make data stationary, each term will be subtracted from the previous. This process can be repeated multiple times, and the amount of times the data is differenced is denoted as $d$. Therefore, our final model will be written as ARIMA($p$, $d$, $q$), where p, d, and q are the orders of the respective terms \cite{Hyndman}.
    
    The SARIMA model is similar to the ARIMA, except it takes into account the seasonality of the data. It assumes that a data point at time $t$ is similar to the data point at time $t - m$ for some predetermined $m$ based on domain knowledge. For example, for an hourly data set $m$ would be 24 if the data exhibited daily patterns. The AR, MA, and differencing are all done on the data points in previous seasons as well as the data points immediately preceding the predicted point. Therefore, a SARIMA model will be written with seven parameters as SARIMA($p,d,q,P,D,Q$)$m$, where $p,d,q$ are the order of the non-seasonal component and $P,D,Q$ are the orders of the seasonal component, and $m$ is the length of the seasons in terms of number of datapoints \cite{Hyndman}. We chose to use a SARIMA model because of the seasonality observed over days and weeks in the underlying data. 
    
    We used the \texttt{statsmodel} implementation of the SARIMA model, which would allow us to learn coefficients using MLE \cite{statsmodels}. Furthermore, we used the python module \texttt{pmdarima}, which is a wrapper that allows us to auto-estimate the model parameters $p,d,q,P,D$ and $Q$ \cite{pmdarima}.
\subsection{Long Short Term Memory Model}
    There has been much historical work in creating models that imitate the human memory behavior of retaining information about previously seen data as new data is considered. Initially, the Recurrent Neural Net (RNN) structure offered a method by which the output of previous predictions were stored and used as an additional input in the next iteration. Unfortunately, in practice these models are ineffective because these dependencies require updates for each prior iteration, leading to either exploding or vanishing gradients \cite{bengio_learning_1994}. This means that data seen outside of the recent history cannot be effectively considered. The LSTM model was proposed to solve this issue. It revolves around a memory cell that is a non-linear combination of an input gate, a forget gate, and an output gate \cite{hochreiter_long-short_1997, cho_learning_2014}. These models were effectively applied for Natural Language Processing task of predicting the next word of a sentence \cite{wang_learning_2016}. The rationale for applying an LSTM for time series data was that this task is similarly linear and depends on trends of previously seen data.
    
    Our implementation of the LSTM was based on PyTorch's \texttt{torch.nn.LSTM} class \cite{noauthor_lstm_nodate}. Each of the previously mentioned gates are defined with the following equations where $i$ refers to the current time step and $h$ refers to the hidden state of the prior time step. Note that the interior of the non linear transforms can be equivalently written with the biases included with the weights when $x=0$ as $Wx_i+Uh$. This clearly demonstrates the different weights that must be trained and how the previous time step's hidden state is used in the current calculation. In each iteration, the model retrains the weights ($W_i$, $W_h$) and biases ($b_i$, $b_h$) for each of the gates.
    
        \[ i = \sigma(W_{ii}x+b_{ii}+W_{hi}h + b_{hi})\]
    The input gate describes to what extent the current time-step influences the prediction.
        \[ f = \sigma(W_{if}x+b_{if}+W_{hf}f+b_{hf})\]
    The forget gate influences how much of the past should be remembered for the current time step
        \[ o = \sigma(W_{io}x+b_{io}+W_{ho}h+b_{ho})\]
    The output gate describes how much of the current time-step should be retained as memory
        \[ \tilde{c} = \text{tanh}(W_{ic}x+b_{ic}+W_{hc}g+b_{hc}\]t
    This equation produces the state of the new memory cell
        \[ c' = f \circ c + i \circ \tilde{c}\]
    The final memory cell can be written as indicated with the forget function applied for the cell's previous state and the input get applied for the new memory cell. The $\circ$ indicates the Hadamard product.
        \[ h_i = o_i \circ \text{tanh}(c')\]
    Afer applying the output gate, the remaining is stored as the time step's final hidden state.

\begin{figure}[hbt!]
    \centering
    \caption{Graphical representation of an LSTM memory cell from \href{https://colah.github.io/posts/2015-08-Understanding-LSTMs/}{Colah}}
    \includegraphics[scale=0.25]{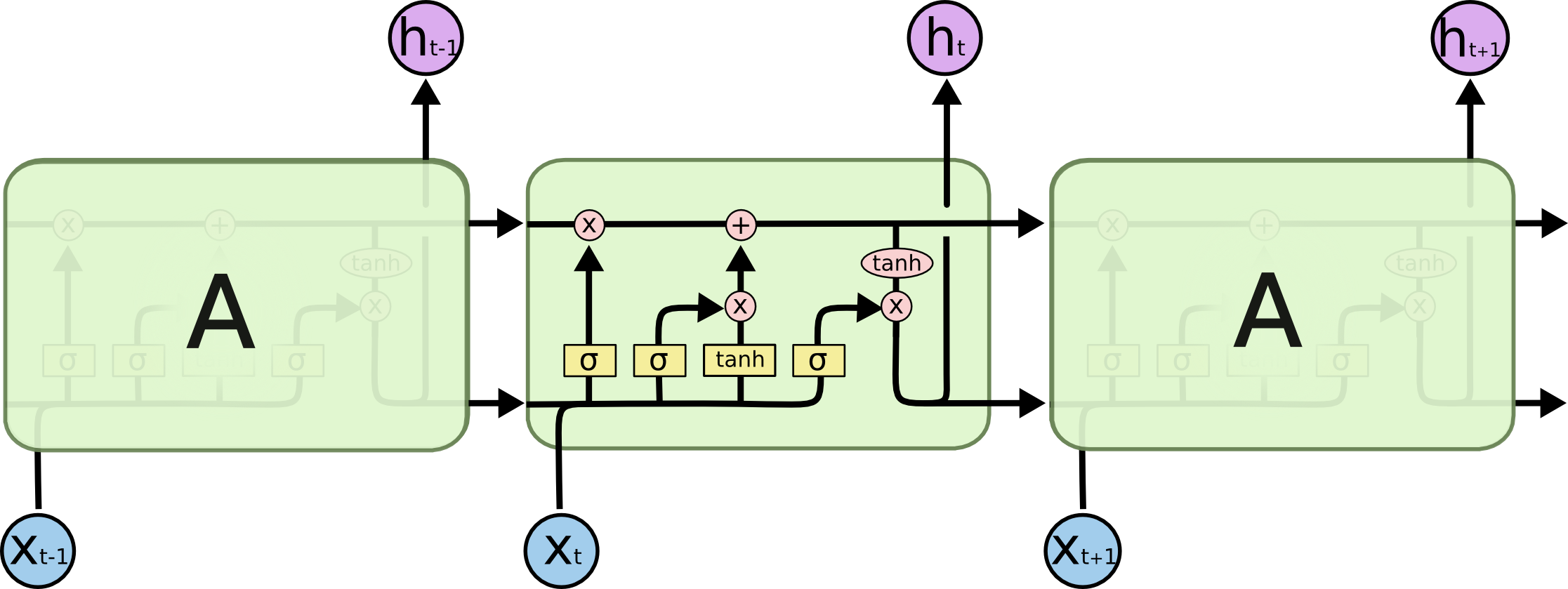}
\end{figure}

\subsection{Evaluation Metrics}
Although there have been many proposed metrics for time-series model evaluation, we chose to focus on the Mean Average Percent Error, Root Mean Squared Error, and the Mean Average Error. For each of the following measures of error, $\hat{y}^t$ denotes the model's forecast, while $y^t$ denotes the actual value at time $t$.

\[ MAPE = \dfrac{1}{n}\sum_{t=1}^n \dfrac{|\hat{y}^t-y^t|}{y^t}\]
\[ RMSE = \sqrt{\dfrac{1}{n}\sum_{t=1}^n (\hat{y}^t-y^t)^2}\]
\[ MAE = \dfrac{1}{n}\sum_{t=1}^n |\hat{y}^t-y^t|\]

The MAPE measurement is widely used in literature and can be interpreted as the percentage difference of the prediction compared to the label. However, it appears that the data sets which have many low or zero-valued scores of 0 have poor performance by MAPE because MAPE approaches infinity as the actual value approaches 0. Therefore, while we report MAPE scores, we also considered the MAE as a more general estimator of the average difference from the prediction and the label. On a data set with possible values from 0 to 100, the worst possible MAE is 100, while a naive model that always 50 would attain an MAE of 50 or less.

\section{Findings}

\begin{figure*}[!t]
    \centering
    \caption{Mean Absolute Error of SARIMA models}
    \includegraphics[scale=0.5]{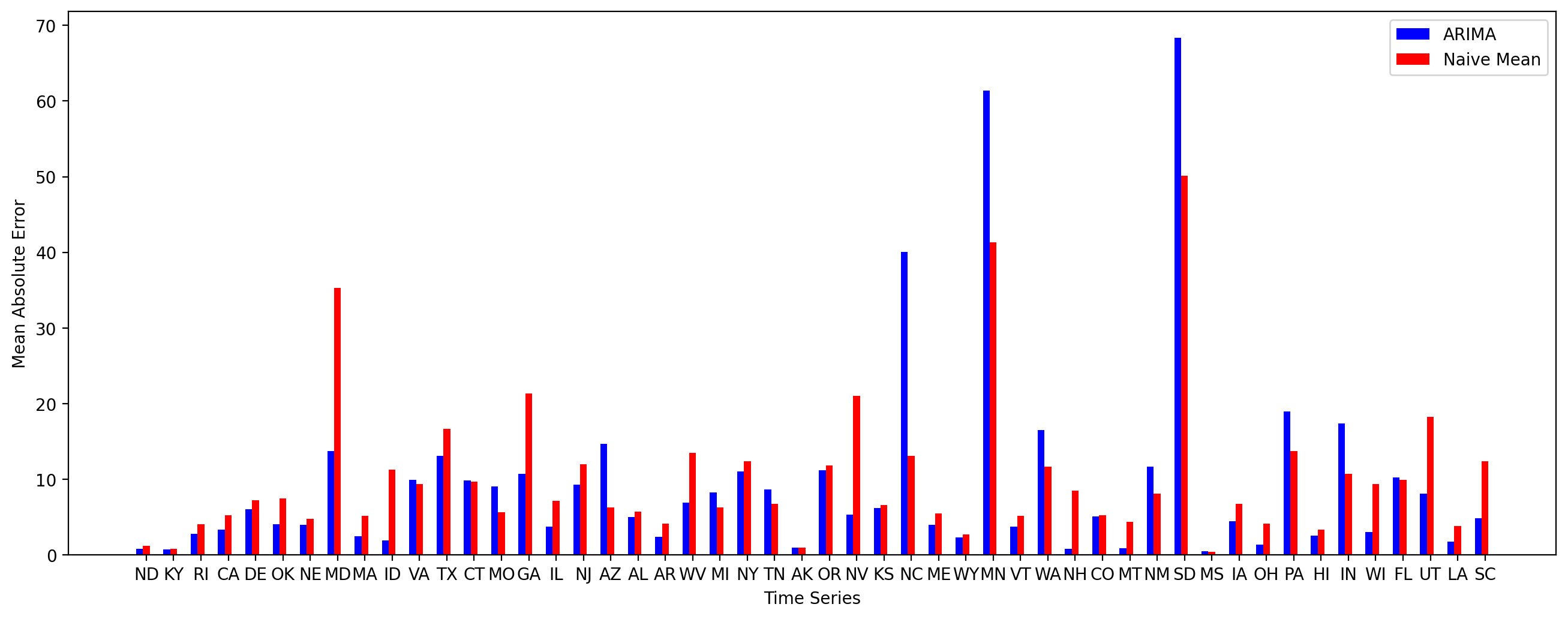}
\end{figure*}
    
 \begin{figure*}[!t]
    \centering
    \caption{Mean Absolute Error of LSTM models}
    \includegraphics[scale=0.5]{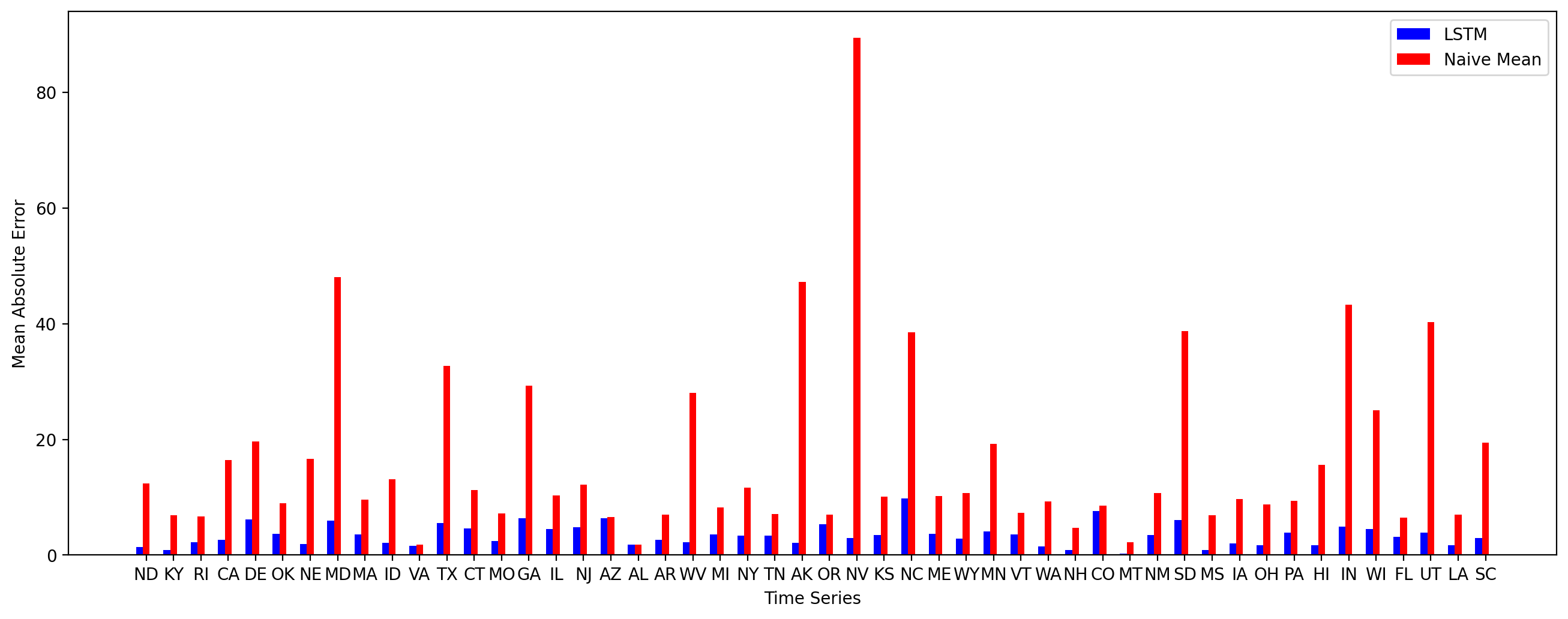}
\end{figure*}

\subsection{SARIMA}
    First, we needed to find effective hyperparameters for our model. Since the data has daily seasons, finding $m$ can be calculated with the following equation, where $s$ is the separation in minutes of each datapoint. 
        \[ m = 24 * (s / 60)\]
    Furthermore, since almost every time series is already stationary (calculated using KPSS), we can use 0 for both $d$ and $D$.
    Unfortunately, the time to fit an SARIMA model with \texttt{statsmodel} increases significantly as $m$ increases. For $s = 20$ it takes almost 40 minutes to train for only one set of hyper parameters. Therefore, for hyper parameter tuning, we downsampled the data into one hour buckets. 
    
    To calculate $p, q, P, Q$ we did a grid search of values from $0$ to $3$ for each parameter using pmdarima's auto arima function. We optimized for the Akaike information criterion, an estimation of out-of-sample error for the time series. We repeated this process for a subset of the data sets. Although the grid search predicted a SARIMA(1, 0, 1, 2, 0, 1)  model, we found empirically that an SARIMA(1, 0, 1, 3, 0, 3) model performed better. Therefore, these hyper parameters were used. 
    
    Next, this model was fit on every training time series at the one hour resolution, then predictions were made for the corresponding training data sets. The MAE for each down sampled data set is shown in Figure 5. Note that 68\% of the time, SARIMA outperforms a naive mean model. 
    
    For our final model, we applied this SARIMA model to the 20 minute data set.  For this set of forecasts, we calculated errors based on both long-term test sets (20\% of the sample) and the short term test set (only an hour long, or three points). Because the SARIMA model had much worse performance on the shorter-interval data sets, it took a significant amount of time (days) to finish training and testing all 50 data sets. A subset of 12 of these results are listed in tables 2 and 3, at the end of this paper. It outperformed the naive mean model for long term prediction, with an average MAE of 8.64 and 10.51 for the SARIMA and naive model respectively. It also outperformed the naive model on the short term prediction horizon, with average MAEs of 5.38 and 6.54. 

    We attempted to do cross validation to increase the sample size. However, this was prohibitively resource expensive, since it would multiply the amount of times we needed to fit the data by two orders of magnitude. Therefore, we make all of our predictions at once, and do not update the values of our time series as we test them. 

\subsection{LSTM}
To effectively tune the LSTM model, extensive hyper parameter tuning was done after implementing a general model. Our hyper parameter search included batch size, max number of epochs, learning rate, hidden layer dimension, number of stacked LSTM layers and dropout. These parameters were tested by selectively scaling a single parameter while fixing the others. Related parameters (such as learning rate and max number of epochs) were considered together. After wide testing, we found reasonable values for batch size (100), max number of epochs (30), and learning rate (0.01) which we standardized to test the remaining parameters. Note that data was loaded using the sliding window method, meaning that each batch included about 35 hours of training or test data. For the sake of this analysis, we considered the average MAE and MAPE measures of error over all the test data of each of the 50 data sets.

In comparison to the naive methods, the LSTM performed exceedingly well. The MAE performance of the Naive and Naive Mean baselines was 16.74 and 16.78 respectively. The MAPE performance of the Naive and Naive Mean baselines was 0.90 and 0.94 respectively. Overall, the fine-tuned LSTM outperformed the Naive Mean model for 84\% of the data sets by the MAPE score and 96\% of the data sets by MAE. Drop out rate seemed to have a considerable impact on model performance with lower dropout rates resulting in better performance. Although with specific parameter changes some models with more layers achieved equivalent performance, models overall performed better with fewer than 4 layers and some of the better results were in models with 1 layer. Models with 10-20 hidden units tended to perform best. The best LSTM model with respect to the measure of MAPE was one with 2 layers, 20 hidden parameters and a drop out of 0 had an MAE of 3.45, but a MAPE of 0.60. Many models performed similarly based on MAPE. In particular the model of 1 layer and 20 hidden units with a drop out of 0.5 performed well on both metrics. A selection of the models tested can be seen in Table 4.

\section{Discussion}
\subsection{SARIMA}
The SARIMA model performed better than the naive methods, but not by an overwhelming amount. Across the whole dataset, 68\% of the time, it had a lower MAE than the naive method. It significantly outperformed the naive mean approach on the long term 3 day test set, and slightly outperformed on the short term test set. Furthermore, we notice that the performance of both the SARIMA and naive models decreased as the variance in the underlying data increased. 

However, there is a huge difference in the performance depending on data set. For example, examine the forecast on the WI data set. 

\begin{figure}[!h]
        \centering
        \caption{SARIMA CPU Forecast for WI Time Series}
        \includegraphics[scale=0.5]{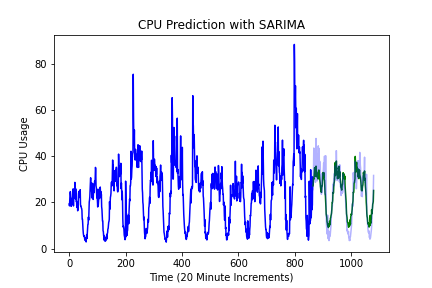}
\end{figure}

We can see that because of the highly seasonal pattern, the SARIMA model was able to perform very well. It had a MAE of  3.65, compared to the naive MAE of 9.83. It was able to closely fit a pattern, despite making the prediction several days into the future.

Now, consider the prediction for the SD time series, one of the worst performing predictions. The SARIMA model had a MAE of 67.75, compared to the naive mean MAE of 50.09. This prediction is actually worse than just predicting a straight line at $y=50$. The SARIMA model searched for patterns that didn't exist, and so ultimately came to a very poor fit. Also, the SARIMA model interprets the on/off nature of the CPU trace as an overall increasing trend in the data, further worsening the model performance. 
\begin{figure}[!h]
        \centering
        \caption{SARIMA CPU Forecast for SD Time Series}
        \includegraphics[scale=0.5]{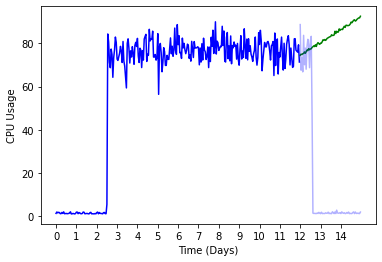}
\end{figure}

Although not shown for brevity, many of the other data sets that performed well were also highly seasonal (such as MD and OK). Furthermore, many of the poor-preforming data sets were had an on/off (VA, MO), erratic burst (AL), or highly noisy pattern.  Therefore, we can conclude that the SARIMA model is very effective for seasonal data, but very ineffective for data that displays erratic patterns, random bursts, or an on/off pattern. By using time series decomposition and prior knowledge of workloads, it could be possible to decide in advance whether SARIMA should be used.

\begin{figure}[!b]
        \centering
        \caption{SARIMA CPU Forecast for MD Time Series}
        \includegraphics[scale=0.5]{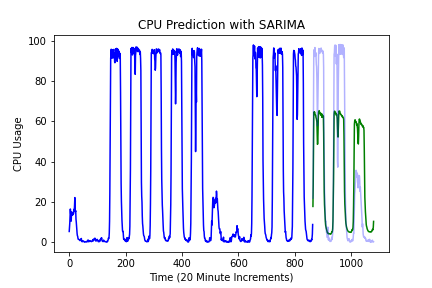}
\end{figure}

Another shortcoming of the SARIMA model used is that it can only make predictions based on one type of season, in our case, the daily patterns in CPU usage. It could not predict long term patterns in the data, such as weekly patterns. For example, the MD data set exhibits very strong usage on weekdays, and very low usage on weekends. The SARIMA model averaged these two different types of days, meaning that weekdays were underpredicted and weekends were over predicted. A more general model could overcome these differences. However, it's still important to note that despite this problem, the SARIMA model was still twice as accurate as the naive model for the MD data set (despite the data sets very high variance). 

\subsection{LSTM}
The PyTorch implementation of the LSTM was very well optimized, especially for use with a GPU. As a result, the model could be efficiently and tested on the entire set of 21600 points in a reasonable amount of time. In this task, the model would have an input of 120 points representing the prior 120 minutes and would be expected to output 60 points for the following 60 minutes of data. Without significant hyper parameter tuning, the model performed better than the naive methods by MAE and marginally worse by MAPE. It had an average MAE of 2.58 and an average MAPE of 1.05 In order to compare the results of the model with the SARIMA model, we down sampled the data which resulted in the model needing to predict the next 3 points with the previous 6. The model tended to perform much better on the down sampled data. This is most likely because in down sampling our data, we took the average of every 20 points, resulting in the smoothing of the data. Whereas the original data was more varied and fluctuated frequently, this averaged data was more smoother which lends itself to easier prediction.

\begin{figure}[h]
        \centering
        \caption{LSTM CPU Forecast for OR Time Series}
        \includegraphics[scale=0.5]{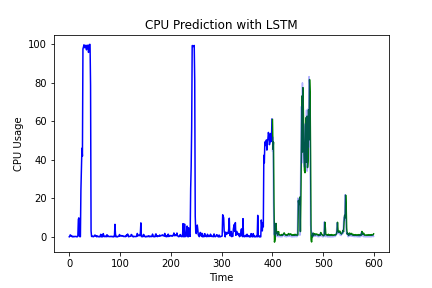}
\end{figure}

After considering more specifically each of the data sets and their respective performance, it seems that the MAPE measure does not necessarily give meaningful information about the performance of the model. Note that Figures x, y, and z are the representations of the first of the three points the LSTM model predicted at each sliding window test data point. Consider Figure 10 that displays the prediction for the OR data set. Although it appears visually that the model is well fit to this data, the MAPE score for this data was 1.42. This high score indicates that, on average, the model made a prediction 142\% above or below the correct label. In contrast, the MAE score was more reasonable at 5.18. This score indicates that on average, the model was 5.18 units (out of 50 max) above or below the correct label. This is discrepancy is because in the calculation of MAPE, the percent is calculated with the label as the denominator. As a result, if the denominator is 0 or close to 0, the MAPE has a higher likelihood of being very high. Notice that in the OR test set, many of the actual CPU usage values are 0 or very close to 0. This trend was further observed in other similar data sets, such as SD and AL.

It is possible that some of the data sets posses similar characteristics and patterns, meaning that training on a subset of them will result in models that have better performance on those data set groups. The underlying assumption here is that there is an unobserved pattern between how clusters are being used. In the course of this research, we attempted to build a multivariate LSTM model that considered the 50 data sets concurrently at each time step. For consistency with our other models, the input vector had 6 time steps of data (two hours) and the model had an output of a 3 time step prediction (one hour). Although the LSTM is capable of receiving multidimensional data, it can only use this information to predict a single variable at a time of multiple dimensions. This means that for each bach size, the input vector was of size \texttt{(100, 6, 50)} and the output vector was of size \texttt{(100, 3)}. The hyper parameters were used of the model that performed best on the non-concurrent data and the results were average. The model had a mean MAE of 10.61 and a MAPE of 1.78. This work is worthy of its own extensive parameter search and study and may be promising in future research of this time-series data.

\begin{figure}[h]
        \centering
        \caption{LSTM CPU Forecast for SD Time Series}
        \includegraphics[scale=0.5]{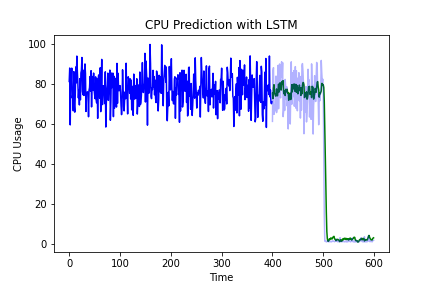}
\end{figure}

\begin{figure}[h]
        \centering
        \caption{LSTM CPU Forecast for WA Time Series}
        \includegraphics[scale=0.5]{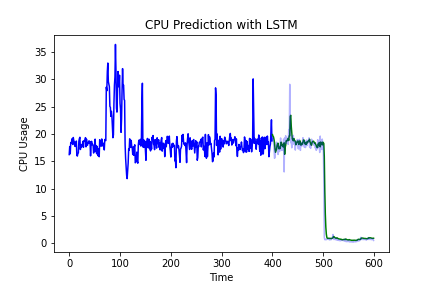}
\end{figure}
\subsection{LSTM and SARIMA Comparison}

\begin{table}[H]
\centering
\resizebox{\columnwidth}{!}{
\begin{tabular}{ccccc}
\hline
     Model    & S.T. MAE & L.T. MAE & S.T. MAPE & L.T. MAPE \\
        \hline
SARIMA   & 5.38          &  8.64            &  0.67        & 5.29          \\
LSTM    & 3.49           &  10.6            &  0.49        & 1.96          \\
Mean    & 9.73           &  10.51           &  1.58        & 5.76             \\
\hline
\end{tabular}      
}
\caption{\label{tab:Table 2}Average Accuracy Metrics of LSTM and SARIMA models}
\end{table}

    In table 1, we can see the average MAE and MAPE of all the time series tested, across both long and short term ranges. We can see the LSTM generally outperforms the SARIMA model for short term forecasting. However, the SARIMA performs better in the long-term, based on MAE. The SARIMA model outperformed the naive ones on each task for each metric. The LSTM model performed better than the naive models for the short term task, but marginally worse on the long term with respect to MAE.
    
    The SARIMA model is heavily based on the daily seasonal pattern of CPU usage. Therefore, it makes sense that it performs well under long term conditions, because it can leverage the highly seasonal patterns in the data. In the process of fitting the data, it considers all of the input data at once, in one data set. This allows it to establish long-term trends. However, the LSTM model takes in a limited data set at each iteration, limited by the batch size. In particular, we used a hyper parameter batch size of 100 which means that approximately 35 hours of data was considered in each iteration. This means that the model is limited in forming dependencies larger than 35 hours. This reduced input size logically lead to worse performance. The SARIMA model, on the other hand, trains the model the same way no matter how many points are predicted. 
    
    The LSTM model vastly outperformed the SARIMA model in terms of computation.  To train the LSTM model for one time series took on the order of 30 seconds. By contrast, the SARIMA model took 45 minutes to train for a time series of 20 minute intervals. This is because the SARIMA model suffers extremely bad performance as the time interval of the time series decreases. This could be a function of the statsmodel implementation and not of the underlying model, however (the R model allegedly has significantly better performance). 
    
    However, based on figure 6, we can see that LSTM outperforms random error in 96\% of cases, whereas SARIMA outperforms the naive model of only 68\%. This is because SARIMA heavily depends on its modeling assumptions being correct (specifically, that the data is seasonal). SARIMA can also respond relatively badly if the data is on/off. In contrast, LSTM makes less assumptions about the underlying data structure, and therefore, performs better on eccentric data sets. For example, consider the SD time series (Figure 8, Figure 10). The LSTM could easily handle the sudden drop off in CPU usage, whereas the SARIMA model would not.

\section{Conclusion}

Although the SARIMA model performed worse for the short term evaluation, it performed better for the long term trend prediction. It is possible that the model may perform even better with more resources for fitting, a brute force hyper parameter optimization and the use of all of the data without down sampling. Although it seems infeasible to repeatedly calculate predictions from the SARIMA model, it may be effectively used to gauge performance in the long term. Previous work in cluster resources analysis highlights the importance of predicting usage long term to estimate how resource demand will grow or shrink \cite{kumar_forecasting_2016}.

As the problem was outlined to us by the Microsoft Azure researcher, their primary objective is to predict the highest anticipated CPU usage over the next hour and scale up (or scale down) their clusters by a fixed amount to reach that level of demand. The decision to scale up or down is driven by a calculation of risk that considers the relatively costs of idle cluster resources when compared to needing to rapidly increase resources to meet sudden demand as well as the time it takes to allocate and deallocate resources. For this task, the algorithm better suited to making repeated predictions for the next one hour interval will be more effective. As shown in our comparison of the models, the LSTM model can more effectively predict the future CPU usage for the next hour from both a performance and accuracy comparison. This research serves as a necessarily baseline for the eventual implementation for cluster resource prediction.

\section{Future Work}

We can continue to refine the SARIMA model. Auto-parameter tuning can lead to potential problems like overfitting. We also hope to keep exploring the seasonality aspects of the data. We also would like to explore other classical statistical models, like exponential models. Finally, we will try to find a way to run the ARIMA model without downsampling the data (which was necessary because of computation restrictions). This will also allow for a more accurate comparison between ARIMA and LSTM. 

There is more work we can do in understanding the trends being established by the LSTM models. By examining the trained variables of the LSTM, we can seek to understand how the data is algorithmically being interpreted to gain a greater understanding of the patterns in the data. This could inform a more reliable equation for predicted usage. To improve the model's performance, we can also attempt to run a clustering algorithm on the individual data sets to group them by usage patterns. Based on these clusters, we could train each group of similar data sets with the same LSTM model. This work would be based on the prior belief that there is an underlying similarity between how certain clusters are used. Because the clusters collect data simultaneously, the input vectors can be the concurrent measurements for each group of clusters. This model architecture was briefly described and implemented, but could be made more thorough by the implementation of different models for different data sets. Though this ensemble method, each cluster can be more reliably sorted and predicted.

Transformers \cite{vaswani_attention_2017} and their self-attention layers made a huge impact on the field of Natural Language Processing and have very recently even begun to be applied to Image Classification \cite{carion_end--end_2020}. It is possible that the transformer architecture could be effectively applied to time series data. This area of work is relatively new in time-series analysis and may provide new performance improvements much like it has for NLP.

Finally, we need to be able to not just predict CPU usage, but translate this prediction into appropriate recommendations to allocate cloud resources (in this context, cloud VMs). To deploy these models, they must not only predict the appropriate amount of VMs, but also account for costs in adding or freeing VMs. We will need to determine an appropriate loss heuristic based on the amount of applications running low on compute power, versus wasted compute power. It is possible that in the application of these models for those tasks, there is value in being able to make more granular predictions over minutes rather than 20 minutes at a time. In that case, further model analysis and training must be done to make minute predictions.

\section{Acknowledgements}
This research was initiated as the work for Stanford University's CS 229 Machine Learning course. Course instructors and teaching staff members were important mentors for this project. Additionally, the data was obtained from a researcher working on Azure compute at Microsoft. After research was complete, the findings of this work was shared with the Microsoft team for feedback.

\bibliographystyle{abbrv} 
\bibliography{ms}  

\afterpage{
\clearpage
\Huge{\bf{Appendix}}
\begin{table}[b!]
\centering
\resizebox{\columnwidth}{!}{
\begin{tabular}{lllll}
data set & MAE   & MAPE & Naive MAE & Naive MAPE \\
\hline
IA      & 5.05  & 0.37 & 6.98      & 0.87       \\
VT      & 4.97  & 2.37 & 5.58      & 2.57       \\
WY      & 2.95  & 0.36 & 3.20      & 0.41       \\
NC      & 13.27 & 1.85 & 13.51     & 2.00       \\
WV      & 7.62  & 0.51 & 13.81     & 1.09       \\
ID      & 2.07  & 0.44 & 11.57     & 4.86       \\
MD      & 19.12 & 3.24 & 35.64     & 16.50      \\
NE      & 3.75  & 0.29 & 5.01      & 0.32       \\
OK      & 6.27  & 4.91 & 7.61      & 5.48       \\
DE      & 7.53  & 0.99 & 8.19      & 1.01       \\
CA      & 5.24  & 0.60 & 5.48      & 0.58       \\
ND      & 1.11  & 0.12 & 1.41      & 0.14       \\
KY      & 0.98  & 0.13 & 0.98      & 0.13       \\
WI      & 3.65  & 0.29 & 9.83      & 0.88       \\
\hline
Average & 5.97  & 1.18 & 9.20      & 2.63       \\
Maximum & 19.12 & 4.91 & 35.64     & 16.50      \\
Minimum & 0.98  & 0.12 & 0.98      & 0.13       \\
\hline
\end{tabular}
}
\caption{\label{tab:Table 1} Long term performance of ARIMA model (3 day test set). Note: a subset of 12 of the 50 data sets were included to give a limited representation of the variability in the datasets.}
\end{table}

\begin{table}[b!]
\centering
\resizebox{\columnwidth}{!}{
\begin{tabular}{lllll}
data set & MAE   & MAPE & Naive MAE & Naive MAPE \\
\hline
IA      & 2.24    & 0.31     & 0.53         & 0.08          \\
VT      & 2.81    & 0.65     & 3.84         & 1.93          \\
WY      & 2.99    & 0.20     & 3.02         & 0.21          \\
NC      & 12.62   & 1.87     & 18.93        & 2.88          \\
WV      & 1.08    & 0.05     & 3.09         & 0.16          \\
ID      & 3.67    & 0.61     & 6.67         & 4.48          \\
MD      & 10.71   & 0.24     & 15.14        & 0.38          \\
NE      & 2.82    & 0.19     & 3.23         & 0.22          \\
OK      & 4.19    & 0.86     & 4.41         & 2.27          \\
DE      & 3.93    & 0.98     & 9.86         & 2.26          \\
CA      & 2.49    & 0.22     & 2.73         & 0.25          \\
ND      & 4.56    & 0.21     & 4.54         & 0.21          \\
KY      & 0.46    & 0.07     & 0.55         & 0.09          \\
WI      & 1.89    & 0.06     & 5.34         & 0.17          \\
\hline
Average & 4.03    & 0.47     & 5.85         & 1.11          \\
Minimum & 12.62   & 1.87     & 18.93        & 4.48          \\
Maximum & 0.46    & 0.05     & 0.53         & 0.08          \\
\hline
\end{tabular}
}
\caption{\label{tab:Table 2} Short Term performance of ARIMA model (1 hour test set). Note: a subset of 12 of the 50 data sets were included to give a limited representation of the variability in the datasets.}
\end{table}

\begin{table}[!htb]
\centering
\resizebox{\columnwidth}{!}{
\begin{tabular}{ccccc}
\hline
                     Layers & Drop out &  Hidden Size &           MAE &        MAPE \\
\hline
                          1 &        1 &            2 &          4.14 &        0.76 \\
                          1 &        1 &            5 &          3.54 &        0.51 \\
                          1 &        1 &           10 &          3.50 &\textbf{0.49} \\
                          1 &        1 &           20 &  \textbf{3.48} &\textbf{0.49} \\
                          2 &        0 &            2 &          4.61 &        0.77 \\
                          2 &        0 &            5 &          3.65 &        0.52 \\
                          2 &        0 &           10 &          3.48 &        0.51 \\
                          2 &        0 &           20 &  \textbf{3.45} &        0.60 \\
                          2 &      0.5 &            2 &          5.25 &        1.29 \\
                          2 &      0.5 &            5 &          3.85 &        0.56 \\
                          2 &      0.5 &           10 &          3.63 &        0.52 \\
                          2 &      0.5 &           20 &          3.69 &\textbf{0.49} \\
                          2 &        1 &            2 &         11.05 &        2.59 \\
                          2 &        1 &            5 &         11.03 &        2.69 \\
                          2 &        1 &           10 &         11.12 &        2.65 \\
                          2 &        1 &           20 &         11.06 &        2.69 \\
                          3 &        0 &            2 &          6.36 &        1.03 \\
                          3 &        0 &            5 &          3.73 &        0.52 \\
                          3 &        0 &           10 &          3.63 &\textbf{0.48} \\
                          3 &        0 &           20 &  \textbf{3.58} &        0.48 \\
                          3 &      0.5 &            2 &          6.83 &        2.17 \\
                          3 &      0.5 &            5 &          4.10 &        0.70 \\
                          3 &      0.5 &           10 &          3.82 &        0.57 \\
                          3 &      0.5 &           20 &          3.70 &        0.58 \\
                          3 &        1 &            2 &         11.06 &        2.66 \\
                          3 &        1 &            5 &         11.17 &        2.64 \\
                          3 &        1 &           10 &         11.08 &        2.68 \\
                          3 &        1 &           20 &         11.07 &        2.66 \\
\hline
\end{tabular}
}
\caption{\label{tab:Table 3}Evaluative performance of varying hidden size and drop out for a model with 2 layers. MAE and MAPE refer to the average errors over the 50 data sets. Bolded numbers indicate the lowest values for MAE and MAPE among the other models with the same number of layers. Note that a single layer network may not have a dropout rate other than 1.}
\end{table}
\clearpage
} 

\afterpage{
\clearpage
\begin{figure}[!t]
        \centering
        \caption{LSTM MAPE Performance}
        \includegraphics[scale=0.5]{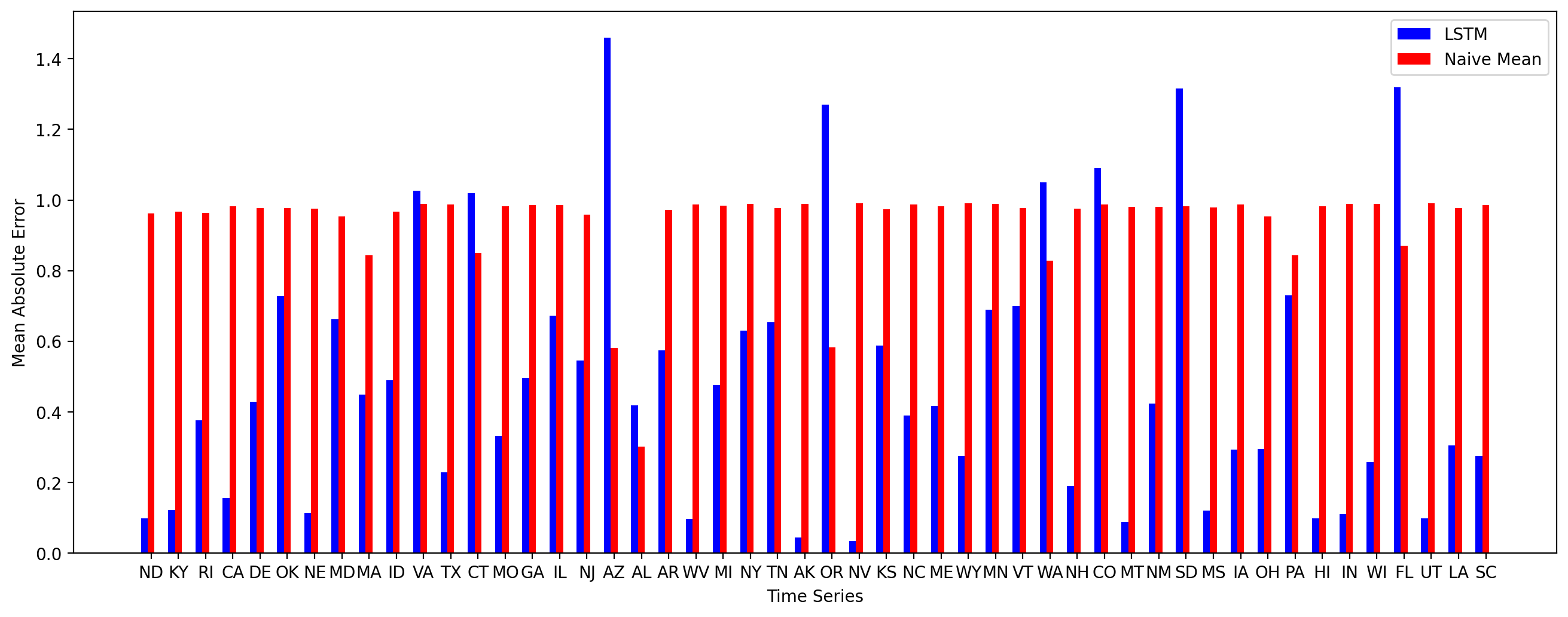}
\end{figure}

\begin{figure}[!t]
        \centering
        \caption{LSTM Concurrent Model MAPE Performance}
        \includegraphics[scale=0.5]{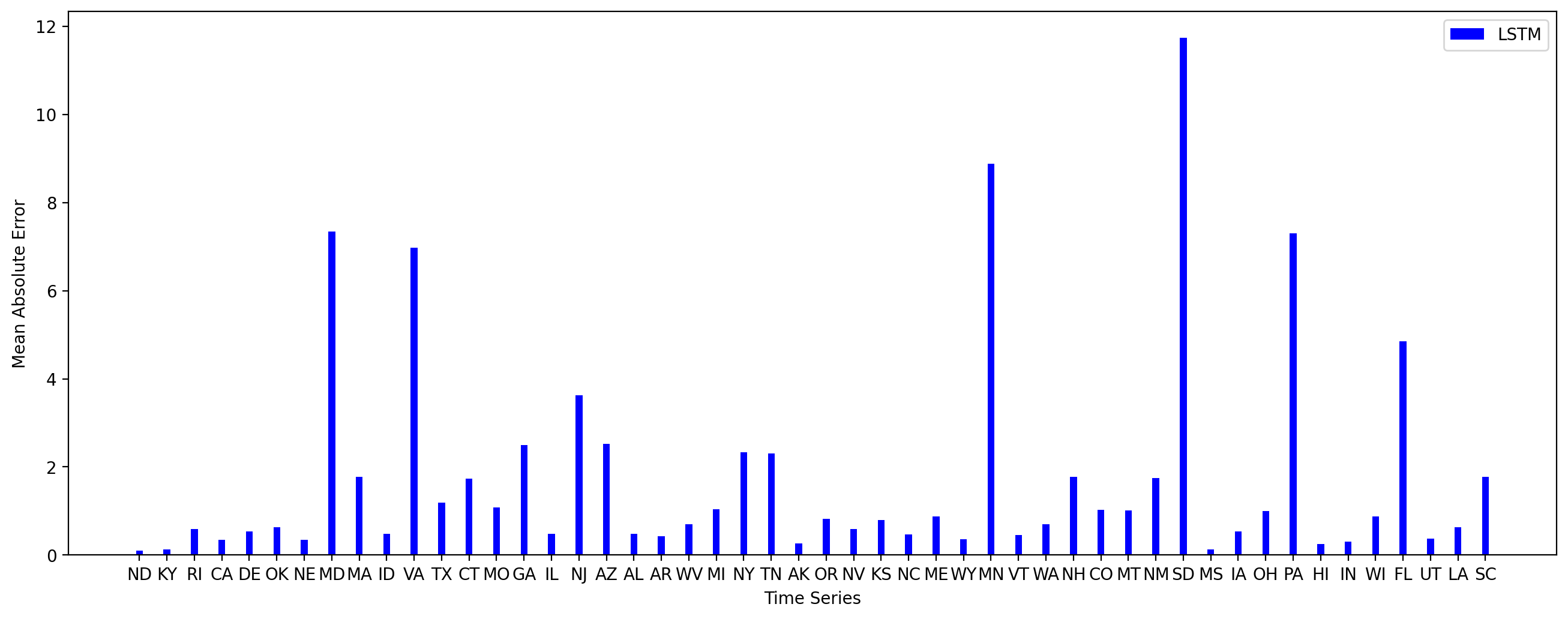}
\end{figure}

\begin{figure}[!t]
        \centering
        \caption{SARIMA Model Short Term Performance}
        \includegraphics[scale=0.5]{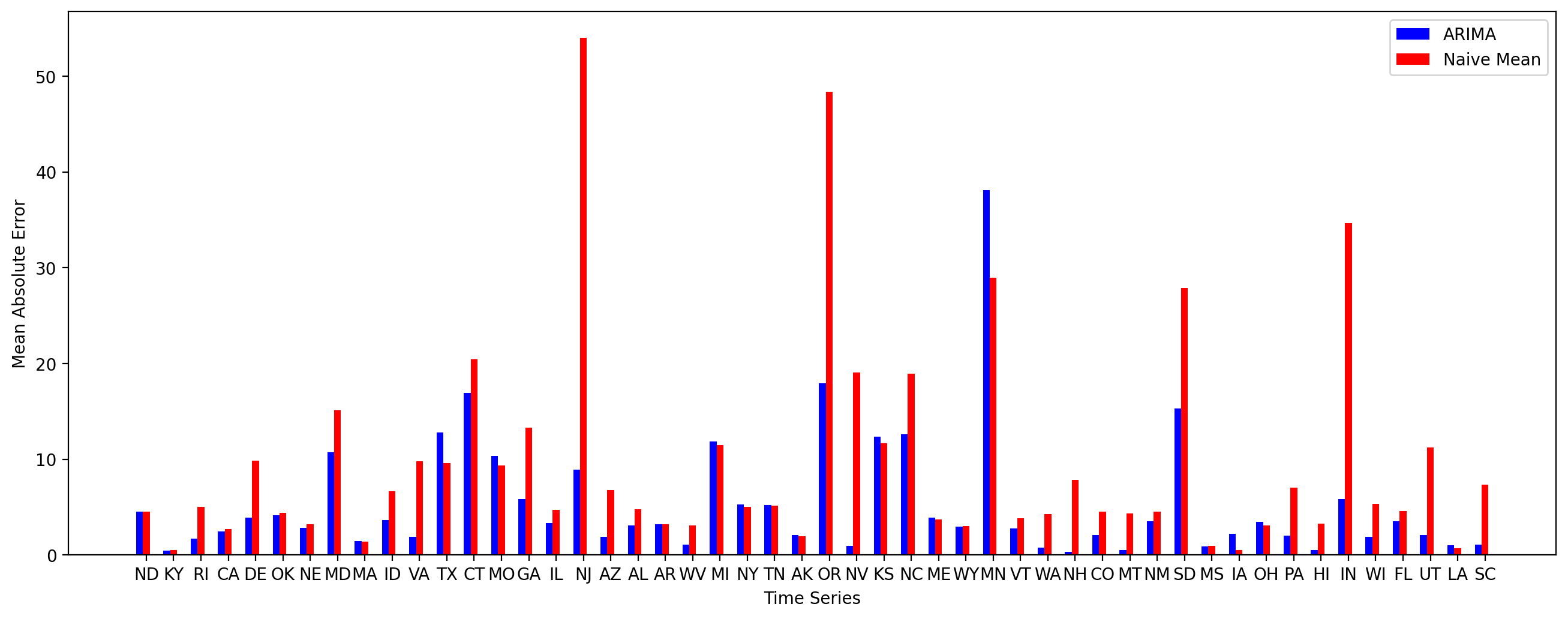}
\end{figure}
\clearpage
} 

\balancecolumns
\end{document}